\pdfoutput=1

\documentclass[11pt]{article}

\usepackage[final]{coling}

\usepackage{times}
\usepackage{latexsym}

\usepackage[T1]{fontenc}

\usepackage[utf8]{inputenc}

\usepackage{microtype}

\usepackage{inconsolata}

\usepackage{graphicx}

\usepackage{todonotes}
\usepackage{booktabs}
\usepackage{algorithm}
\usepackage{algpseudocode}
\usepackage{fancyvrb}

\title{Hands-off Image Editing:\\
Language-guided Editing without any Task-specific\\ Labeling, Masking or even Training} 

\author{Rodrigo Santos \and António Branco\and João Silva  \and João Rodrigues \\
        University of Lisbon\\
    NLX---Natural Language and Speech Group, Department of Informatics\\
    Faculdade de Ciências, Campo Grande, 1749-016 Lisboa, Portugal\\
    \texttt{\{rsdsantos, antonio.branco, jrsilva, jarodrigues\}@fc.ul.pt}}

\begin{document}
\maketitle
\begin{abstract}
Instruction-guided image editing consists in taking an image and an instruction and delivering that image altered according to that instruction. State-of-the-art approaches to this task suffer from the typical scaling up and domain adaptation hindrances related to supervision as they eventually resort to some kind of task-specific labelling, masking or training. We propose a novel approach that does without any such task-specific supervision and offers thus a better potential for improvement. Its assessment demonstrates that it is highly effective, achieving very competitive performance.
\end{abstract}

\section{Introduction}
\label{sec:introduction}

\begin{figure*}[ht]
    \centering
    \includegraphics[width=1\linewidth]{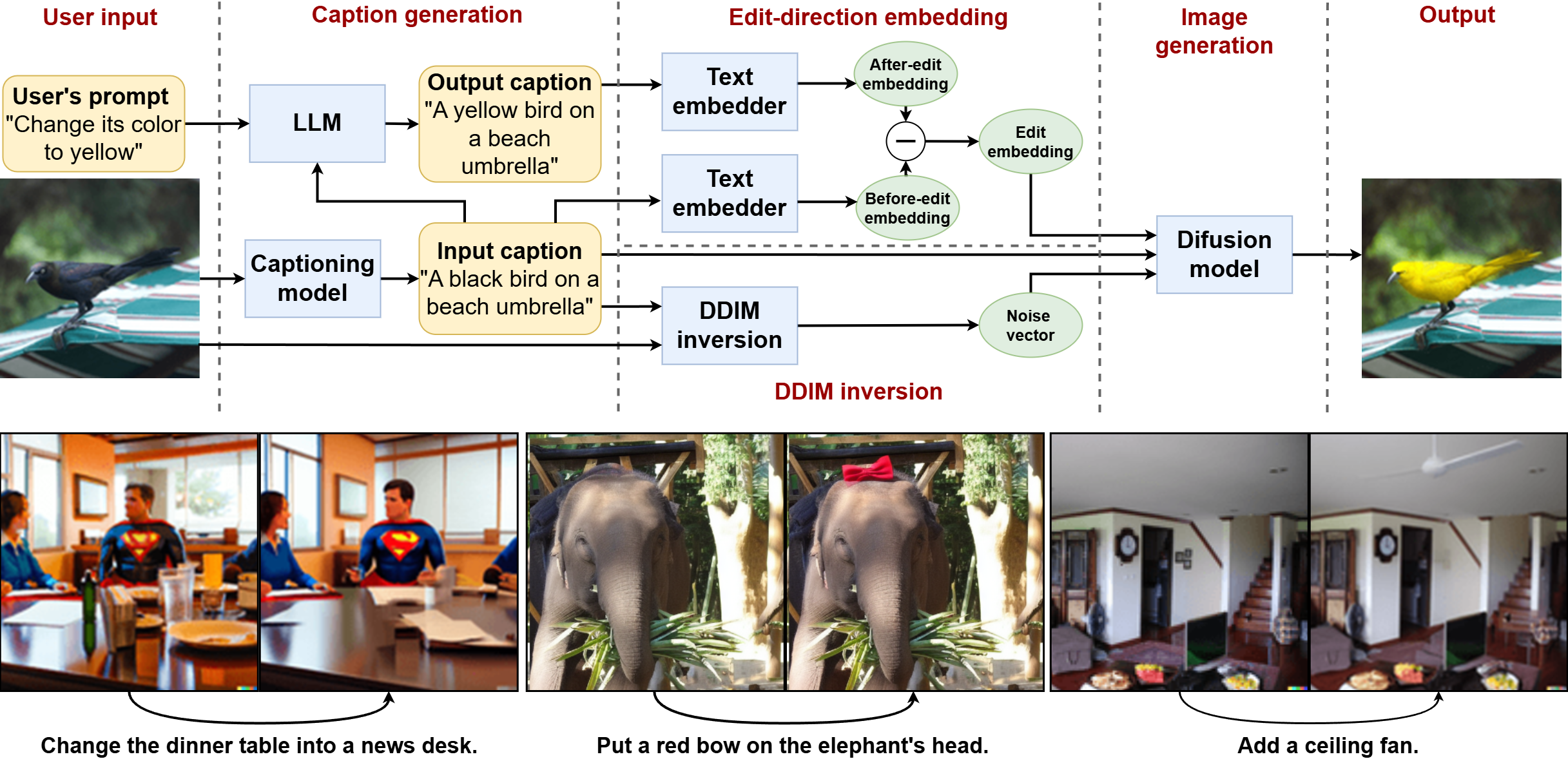}
    \caption{Top: Architecture of the method presented in this paper. Bottom: Three examples, showing inputs (left images and edit requests) from the MAGICBRSUH test set and outputs (right images) from our proposed method.}
    \label{fig:architecture}
\end{figure*}

Instruction-guided image editing consists in taking an image and an instruction in plain natural language stating an intended alteration and delivering another image that results from altering the input image according to that instruction, as in Figure~\ref{fig:architecture}.

The alteration should be confined to the aspects objectively requested in the instruction, keeping the rest of the image unaltered.
Provided they are sufficiently objective, the intended alterations can be of any kind, from shallow aspects (e.g.~color changing) to more structural contents of the input image (e.g.~add, replace or remove elements).
 
State-of-the-art approaches to this task resort to supervised training~\cite{santos:2022:languageDriven,sheynin:2024:emu,fu:2023:MGIE,zhang:2023:hive,brooks:2023:instructpix2pix,li:2024:zone,santos:2022:lxdrim}, which relies on task-specific datasets, curated to contain triples with an image, an editing instruction, and a possible output image~\cite{zhang:2024:magicbrush,hui:2024:hq-data,ge:2024:seed-data}.
 
These approaches may be complemented with further supervision at inference time, for instance with some kind of masking, where the human user marks in the input image the area that should be affected by the alteration stated in the instruction, e.g. by circling \cite{wasserman:2024:inpaint}.
Alternatively, this supplementary information can itself be learned, in which case further task-specific training is needed \cite{kirillov:2023:sam,li:2024:zone}.

This approach involves hindrances of different sorts.
First, asking the user to manually mark the input image is a step that ideally should be dispensed with in favour of a fully self-contained language-guided operation.
 
Second, while requiring extensive and costly labelling effort, task-specific datasets induce the typical bottleneck in terms of domain adaptation, and the supervised training they support suffers from the typical ceiling in terms of scaling up, as larger and more diverse datasets are ever needed to achieve improved results.
 
Finally, task-specific datasets inherently suffer from an insufficient representability of the task they are aimed at capturing as in each instance the output image stored is just one among many equally possible outcomes of the respective instruction over the respective input image.
 
To illustrate this, consider an example where the input image contains an elephant and the instruction asks to place a bow on it (see Figure~\ref{fig:architecture}): producing an image where the elephant is now wearing a small red bow at the top of its head is an acceptable outcome, but it is just one among many possible others, where the bow would appear with different sizes, shapes, colours, in different positions, etc.
 
This insufficient representability is problematic for the performance of the system, which happens to be trained on a rather narrow sample of possible outcomes and thus is hindered in its generalization capacity.
But as these datasets serve not only for training but also for evaluation, this is also problematic for the fair assessment of the systems seeking to solve this task: if they too do not deliver an altered image that is identical or close to the output image stored in the gold test dataset, that counts as a failure, even though that altered image is a different, yet fully acceptable response to the instruction.
 
In this paper we propose an alternative method to the instruction-guided image editing task that dispenses with masking, task-specific datasets, and even any task-specific training (supervised or not), circumventing the hindrances noted above.

Our method is sketched in Figure~\ref{fig:architecture}. The instruction is integrated into a suitably designed prompt.
That prompt is entered into a pre-existing LLM (not trained for this task).
That LLM generates a caption for an input image (but not that image) and for an output image (but not that image).
The embeddings of these two captions are used to obtain a difference vector that represents the alterations required for the input image.
Together with the input image, this edit-direction vector is entered into a diffusion model that generates an output image.

Experimental results reported in the present paper demonstrate that our method is highly effective, achieving competitive performance with regards to state-of-the-art methods, which are much more resource intensive given they rely on task-specific learning and datasets.
 
Also in contrast to those approaches, our method has the potential for its performance to keep improving.
It does not require any training since it only needs pre-trained models (an LLM and an image generation model conditioned on embeddings, such as Stable Diffusion). 
Consequently, as the capacity of these pre-trained models improves pushed by the respective research areas, the performance delivered by our method will also improve.
 
The remainder of this paper is structured as follows:
Section~\ref{sec:rel_work} introduces related work;
Section~\ref{sec:method} presents the proposed approach;
Section~\ref{sec:results} elaborates on the experimental setup;
Section~\ref{experiments} reports on the results obtained and discusses them;
Section~\ref{sec:conclusion} gives concluding remarks;
and finally Section~\ref{sec:limitations} discusses possible limitations of our work.

\section{Background}
\label{sec:rel_work}



Early work focused on GANs and RNNs to perform the editing~\cite{El-nouby:2019:tell,Cheng:2020:sequential,Jiang:2021:language}, demonstrating promising capabilities in both understanding and performing the alterations requested. 
However, these works focus on toy datasets~\cite{El-nouby:2019:tell}, constrained domains~\cite{Cheng:2020:sequential}, or a reduced set of possible edits~\cite{Jiang:2021:language}.  

A more promising approach was advanced by \citet{brooks:2023:instructpix2pix}, who introduced an image editing dataset that is used to fine-tune a Stable Diffusion model.
This is done by giving it a new conditioning vector that represents the input image, and by reusing the same text conditioning mechanism that was originally intended for captions to take as input the text edit instruction instead.

Similarly, \citet{zhang:2023:hive} and \citet{sheynin:2024:emu} also trained a Stable Diffusion model.
The first on a dataset of edited images scored by manual annotation, and the second on a multi-task dataset across a broad spectrum of editing tasks.

\citet{li:2024:zone} propose ZONE, which adds an automatic segmenter, SAM \cite{kirillov:2023:sam}, to select the part of the image that should be edited.
Though the system is claimed to be zero-shot, it includes an InstructPix2Pix module that requires task-specific training.

MGIE \cite{fu:2023:MGIE} goes a step further and incorporates an LLM into the editing pipeline through the use of several adapters and a Stable Diffusion model, all trained specifically for the instruction-guided image editing task.

Though increasingly sophisticated models might bring better performance, their underlying supervised approach runs into intrinsic scalability issues.
They are data-hungry methods that need labelled datasets which are very hard to extend to ever large volumes and more general scope, given the vast space of possible images and instructions.

As an alternative, some proposals sought to resort to unsupervised approaches, like \citet{hertz:2022:prompt2prompt} with a method that involves manual editing of a caption which is then injected into the cross-attention maps during the diffusion process. 

The capabilities of a diffusion model for image editing are also extended by pix2pix-zero~\cite{parmar:2023:pix2pix-zero}, where a vector that helps to guide the model towards the output image is obtained by manually gathering sets of generated captions that support the determination of a difference vector between the input and the output images.

As an example, if the request is to change the hat in a picture into a bow, a large set of sentences about hats is generated with a text decoder, and their latent representations are averaged.
Sentences about bows, in turn, are also generated and their average representation is determined.
The difference vector between these representations is used to guide the generation of the output image.

Nevertheless, these proposals also face intrinsic scalability issues, namely in that the sets of captions require manual curation.
A human in the loop is necessary to understand the image and the editing request, and then to conceive prompts to make the language decoder generate the two sets of sentences that are necessary to guide the image alteration procedure.
Furthermore, the sets of sentences and their corresponding difference vectors are highly specific to each individual request, which limits their ability to generalize.

Differently from previous proposals in the literature, our approach is both unsupervised---overcoming the drawbacks encountered for supervised ones---and guided solely by the editing request---overcoming the drawbacks both of manual assistance to the caption generation process and of its lack of generalization strength.

\section{Proposed approach}
\label{sec:method}

The method we propose relies on a difference vector to guide the image editing, which is obtained from the captions of the input and output images. 
The caption of the input image is obtained with any capable, off-the-shelf image-to-text tool.

The caption of the output image, in turn, is obtained with any capable, off-the-shelf text generation decoder, upon being given the text of the caption of the input image together with the text of the image editing instruction---arranged under an appropriate prompt template.\footnote{ 
This prompt can be generated automatically, with no need for training or manual curation --- see Appendix~\ref{sec:metaprompt}.}
This method is sketched in Figure~\ref{fig:architecture} and is detailed next.

\subsection{Reconstruction after deconstruction}

The generation of the output image consists in the reconstruction of the input image after it has been deconstructed (into its noisy rendering), and it is also done in consonance with the modifications requested in the editing instruction.
This process is known as image inversion and was initially applied to GANs~\cite{Xia:2022:GANInversion,Bermano:2022:StyleGANInversion}. 

To undertake this procedure, here we make use of Stable Diffusion to generate the edited image.
Accordingly, the image is recreated by using the same model, which will allow one to have access to the internal model states during image generation. 

These internal states can then be used together with some conditioning guide~\cite{Abdal:2020:StyleGAN++} to both steer the model to add the requested alterations as well as to ensure that the remainder of the image maintains similarity to its initial form.  

Hence, one of the ingredients needed is the interim noise vector into which the input image is ``deconstructed''. 

To get this noise vector, the inversion process by Denoising Diffusion Implicit Models (DDIM)~\cite{song:2020:DDIM} is resorted to.
This is a technique where the diffusion process is applied in reverse, meaning that one goes from the image to noise, rather than from noise to image. This process requires not only the input image but also a caption of the input image. In our case this caption is obtained with BLIP \cite{li:2022:blip}.

DDIM Inversion is a lossy process, which implies that some details will be already missing before the (re)construction process of the output image even begins.
Thus, by using better inversion methods, the resulting edited image will be better.


\subsection{Guiding the reconstruction of images}

After performing the DDIM inversion, one needs to find a way to guide the image editing process.
To achieve this, an edit-direction embedding is used, which is a vector that points from the input image to the output one.

To obtain this edit-direction embedding, we resort to two textual captions, one that represents the input image, before the edit, and another that represents the output image, after the desired transformation had been applied.

The embeddings for these captions are obtained through a process that involves using an embedding model that delivers its semantic representation in a latent space that is common between language and images.
In the case of Stable Diffusion, the embedding model used is the CLIP model.

These source and target embeddings are both vectors and subtracting them returns another vector.
Similar to classifier-free guidance~\cite{Ho:2022:ClassifierFree}, this vector can be conceptualized as a direction from the initial image, before the edit, to the final image, after the edit. 

By having this embedding, the editing process is guided by using it to steer the image reconstruction towards the desired transformation.
Essentially, it acts as a conditioning factor that helps to guide the image alteration process into the desired direction.

Summing up, one represents input and input images by means of captions, which in turn are represented by embeddings, which in turn are used to obtain the edit-direction embedding, which in turn is used to generate the output image from a ``deconstructed'' representation of the input image.

\subsection{Guiding the generation of captions}

The quality of the edit is affected by the quality of the before and after-edit captions since they are at the root of the edit-direction embedding.


To generate the before-edit caption, we use the one computed for DDIM inversion, which is obtained through BLIP.
For the after-edit caption, we rely on the ever increasing capacity of Large Language Models (LLMs).
This caption is obtained by prompting the decoder with text that includes both the edit instruction introduced by the user and the before-edit caption produced with BLIP.
\footnote{The prompt template used in the experiments below was designed manually by us after some experimentation. For automatically generated prompts see Appendix ~\ref{sec:metaprompt}.}

Accordingly, all models, either for image or language processing, are used only for inference, not requiring any fine-tuning of other sort of training.

\section{Experimental setup}
\label{sec:results}

This section introduces the models used as well as the evaluation data and metrics.

\subsection{Large language models}

The Transformer architecture~\cite{Vaswani:2017:Transformer} has introduced a new paradigm in NLP, 
having leveraged unprecedented results in virtually all sorts of tasks in language understanding, generation, translation, etc.~\cite{Devlin:2018:Bert,Brown:2020:GPT3,Vaswani:2017:Transformer}.

We experimented with a few prominent Transformer models, namely: (i)~\emph{Gemma} \cite{Mesnard:2024:Gemma}, (ii)~\emph{Llama~2}~\cite{touvron:2023:llama2}, (iii)~\emph{Mistral}~\cite{jiang:2023:mistral}, and (iv)~\emph{Phi-2} \cite{gunasekar:2023:phi}.
More details are in Appendix~\ref{sec:appendix_llms}.

\subsection{Image generation model}

Image generation methods encompass a range of techniques aimed at creating realistic or stylized images from scratch or based on existing ones, encompassing generative adversarial networks (GANs)~\cite{Goodfellow:2014:GANs}, variational autoencoders (VAEs)~\cite{Kingma:2013:VAE} and diffusion models \cite{Dhariwal:2021:diffusion,Ramesh:2022:DALLE2,rombach:2022:stable_diffusion}, among others.

For the experiments reported here, two diffusion model were integrated into our image editing pipeline, namely Stable Diffusion 1.4 and 1.5~\cite{rombach:2022:stable_diffusion}. 

As demonstrated in Appendix~\ref{sec:appendix_emb_importance}, Stable Diffusion 1.4 has better performance than 1.5 in our approach, therefore, we will adopt SD 1.4 for most of the experiments in this paper.
Further discussion on the choice and details of Stable Diffusion can be found in Appendix~\ref{sec:appendix_sd}.

\subsection{Evaluation data}
\label{sec:data}

\begin{figure*}[t]
    \centering
    \includegraphics[width=1\linewidth]{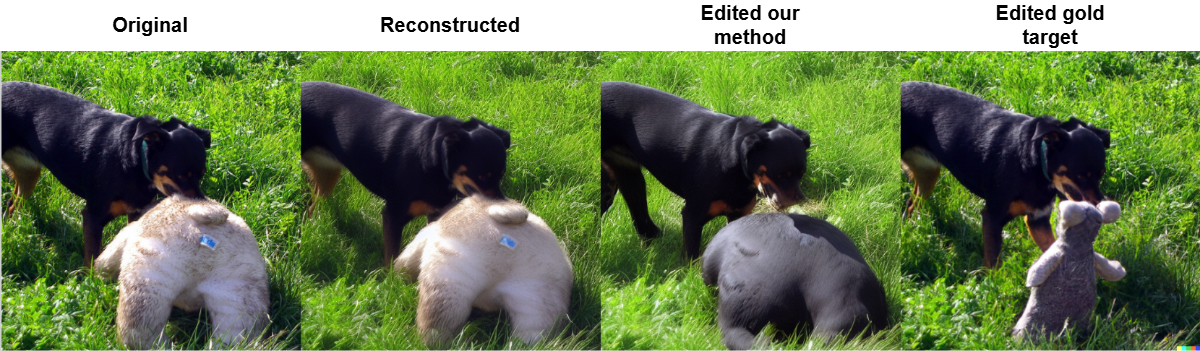}
    \caption{Example from MAGICBRUSH test set.
    The request is ``Make the teddy bear black''.
    The four images are: the original one, the one generated from the noise obtained through DDIM Inversion, the one generated by our system, and the gold edited one in the dataset.}
    \label{fig:dataset_example}
\end{figure*}

For evaluation, we resorted to the MAGICBRUSH dataset~\cite{zhang:2024:magicbrush}.
This is based on images from MSCOCO~\cite{lin:2014:coco} and was created by human curators in Amazon Mechanical Turk with the help of DALL-E~2~\cite{Ramesh:2022:DALLE2}, by having them write a textual edit instruction for each input image, along with the respective target caption for the resulting image, after the intended transformation had been applied, and also drawing a free-form region mask on the input image, indicating the area to be affected by the edit.
The masked image and the target caption were then given to DALL-E~2 to generate a new image with the desired transformation, by using mask infilling.

The dataset includes over 10,000 instances.
Given our model does not require training, we used only the 1053 instances in the test split for evaluation, to ensure comparability with results reported in the literature.

\paragraph{Masking issues}

While, to the best of our knowledge, this dataset is the one with the highest quality for the task, it is not without its problems.
As seen in Figure~\ref{fig:dataset_example}, the teddy bear in the original image (first image) does not have the same shape as the teddy bear in the gold edited image (fourth image), despite the editing request only asking for a colour change (``Make the teddy bear black'').

This problem arises from the masking that was performed on the image during dataset creation, as important information was lost, such as the specific shape of the initial object.
DALL-E~2 only had to generate a new object that fitted the description ``teddy bear'', thus possibly losing relevant aspects of the initial object not transmitted through the textual description.

\subsection{Evaluation metrics}
\label{metrics}

Besides having performed human evaluation, We resorted to automatic metrics, named in Table~\ref{tab:llms} as ``\mbox{CLIP-T}'', ``CLIP-I'', ``BLEU'', and ``Cosine~Sim.'':
\mbox{Except} for BLEU~\cite{Papineni:2002:BLEU}, all metrics are obtained through the cosine distance between CLIP embeddings.

CLIP-I and CLIP-T are obtained by comparing the embedding of the image output by our model to the embedding, respectively, of the gold image (I) and of its gold caption (T).

The other two metrics concern captions, the gold and the generated ones.
Cosine Sim. stands for their cosine similarity and BLEU for the 4-gram BLEU.

CLIP ViT-B/32 was used to obtain all embeddings, which is the same model that was used in the MAGICBRUSH paper, to ensure comparability with the scores reported there~\cite{zhang:2024:magicbrush}.
To compute BLEU, sacremoses was used.\footnote{https://github.com/hplt-project/sacremoses. Default settings were used.}

\subsection{Preferred metric}
\label{sec:prefered_metric}

While comparing to the gold image seems an obvious choice for evaluation, doing so with the gold caption instead may appear as a not-so-useful endeavour. 
Nevertheless, the latter option is not only reasonable but it is even preferable given it permits a more fair performance scoring.

First, as mentioned in Section~\ref{sec:data}, given the way MAGICBRUSH was created, sometimes there is information loss regarding the ``source'' image, such as the shape and colour of entities, especially when these are not specified in the request. 
While this is a problem when using the gold image, in CLIP-I, this issue disappears when using the gold caption, in CLIP-T, because ``a black teddy bear'' is always a description of any black teddy bear despite its specific shape that happens to be represented in the gold image.

Second, DDIM Inversion introduces artefacts and removes or alters details present in the source image (cf.~the first and second images in Figure~\ref{fig:dataset_example} for examples of these differences).
Once again, this problem disappears when using CLIP-T, as details such as the positioning of the blades of grass or the fur pattern on the dog are not important to the distance scoring with it.

Finally, due to the nature of the task, as exemplified in Figure~\ref{fig:dataset_example}, when asked to ``make the teddy bear black'', there are various shades of black that could be used, without making the outcome better or worse.
This is even more prevalent when the request is to add something that was not previously present in the image since often the full details of what is to be added (its shape, colour, position, etc.) are not specified.
When evaluated with the caption, in CLIP-T, unless explicitly specified in the edit request, visual attributes such as shape, colour, position or others do not impact the evaluation outcome---that is not the case with CLIP-I, instead, which turns thus to be a less reliable metric.

\section{Experiments and discussion}
\label{experiments}

This section reports on the assessment experiments. 
The dataset used is MAGICBRUSH, of which eight examples are displayed in Figure~\ref{fig:rand_examples} in Appendix~\ref{sec:appendix_evaluation} to illustrate the output of our system.

Looking through these output images, presented in the second column, one observes very satisfactory results.
All editing requests were made effective at least to some degree, if not completely.

Among all examples where alterations were made, only one clearly does not fulfil the editing instruction, namely the second example on the right, where the ``spider'' was added inside the blender instead of "next to it" as requested.

It is also worth noting the fourth example, on the right, where instead of adding the required object (exotic planet) as a new object in the scene, the model modified a pre-existing object, the traffic light, that happened to have a similar shape/colour.







\subsection{Number of captions and shots}

The approach of \citet{parmar:2023:pix2pix-zero} relies on the generation of “a large bank of diverse sentences for both source s and the target t” (i.e. what we call the before-edit and after-edit captions), whose embeddings are to be averaged.

The specific number of such sentences is not indicated in that paper, but the corresponding code repository mentions “a large number of sentences (\textasciitilde1000)”.

Such a large number of captions is impractical for on-the-fly usage on a consumer-level GPU since generating even a couple of sentences already takes a few seconds. 

To help clarify the number of captions to be used in obtaining the edit-direction embedding, we did preliminary experimentation along these three dimensions:\footnote{After some exploratory experimentation, the prompt used was ``Given the transformation `[TRANSFORMATION]' generate [NUMBER] image captions for before and after the transformation.''---where [TRANSFORMATION] is the textual edit request, and [NUMBER] is the number of before and after-edit captions to be generated.} 
(i)~the number of captions generated, either before-edit or after-edit (one, two and four); 
(ii)~the number of few-shot examples provided to the LLM that generates the caption (zero, one and three);
(iii)~for the before-edit captions, using BLIP, instead of the LLM, to generate the first caption.

Detailed scores from all these experiments are in Appendix~\ref{sec:appendix_n_captions}.

The combinations of these various dimensions supporting the best performance are two: 
(a)~when measured with CLIP-T (0.2817): resorting to 1-shot and 1-caption, and using BLIP for the before-edit caption; 
and (b)~when measured with CLIP-I (0.8350): resorting to 3-shot, 4-captions, and using BLIP for the first before-edit caption.

It is worth noting that using the caption generated by BLIP as the first before-caption permits better performance than having the language model generating it. This can be attributed to the fact that adding the first before-edit caption provides missing information to the model which is not present in the edit request.

\begin{figure*}[tp]
    \centering
    \includegraphics[width=0.9\linewidth]{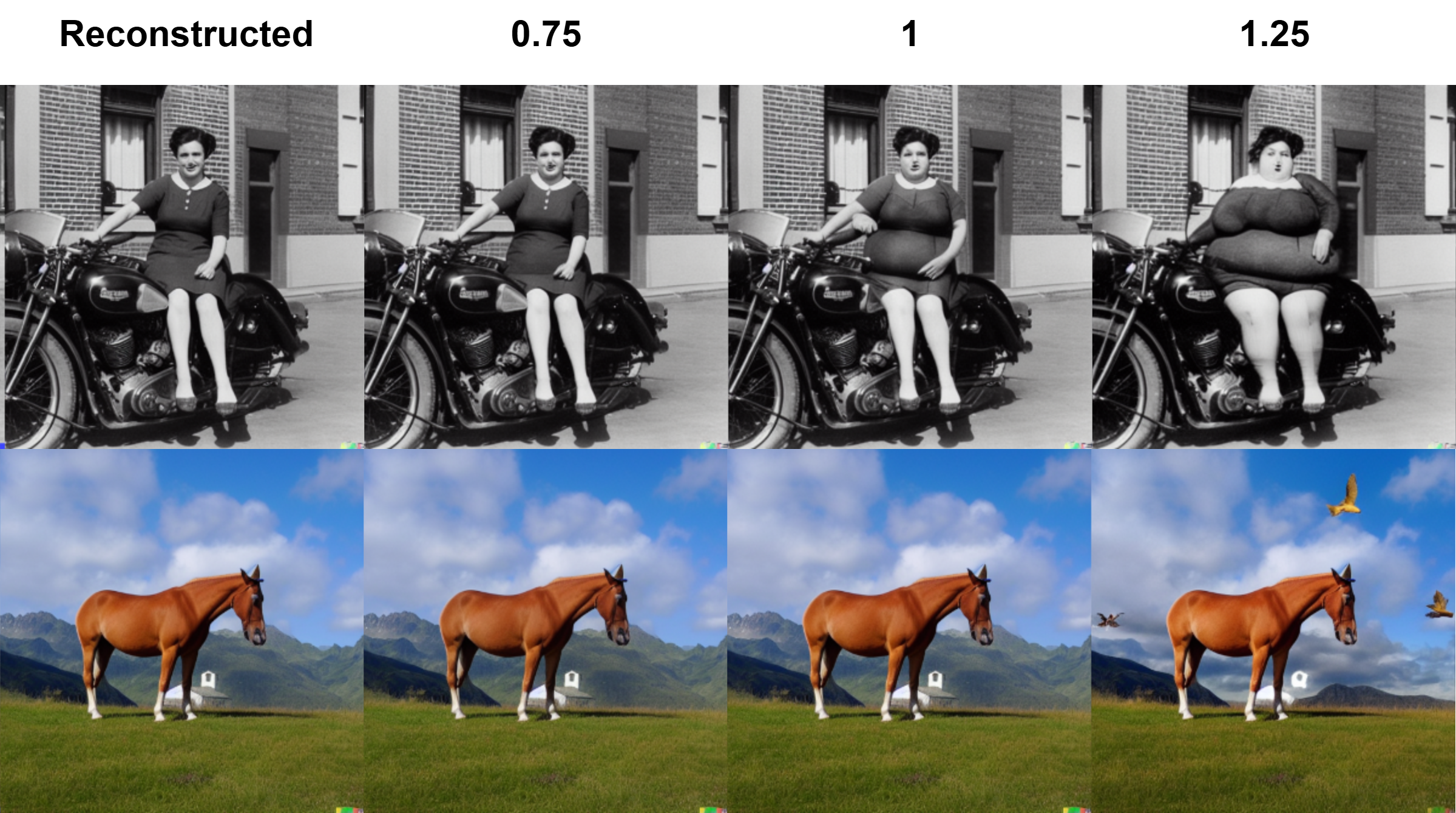}
    \caption{Examples of different edit-direction weights, with base images and instructions from MAGICBRUSH. Instruction in top row: ``Make the woman obese.''; in bottom row: ``Let's add birds to the sky''. }
    \label{fig:examples_importance}
\end{figure*}

\subsection{Prompt simplification}


Under the preferred metric CLIP-T, the best performance is thus found when only one caption is generated for both the before and after-edit captions, and when the before-edit caption is obtained with BLIP. As a consequence, only the after-edit caption is being generated by the LLM.

Considering this, the prompt that asks the LLM to produce the captions was simplified,
and to further help the LLM, we also include in this prompt the before-caption, as generated by BLIP: ``Given the caption `[CAPTION]' describing an image and a transformation `[TRANSFORMATION]' to be applied to the image, generate the caption of the image after applying the transformation.''---where [CAPTION] is the before-edit caption, and [TRANSFORMATION] is the textual edit request.

Using this simplified prompt, there was an improvement with CLIP-T from 0.2817 to 0.2841, and an improvement of 0.8310 to 0.8366 with CLIP-I.

\subsection{Edit-embedding weight}

The edit-direction embedding results from the difference between the embeddings of the before and after-editing captions.
Its importance was assessed by introducing different weighting factors, 0.75, 1 and 1.25, to help find the best edit strength.

A heavier-weighted edit-direction embedding induces the model to make more changes.
This is as expected, and is perspicuously illustrated with the examples in Figure~\ref{fig:examples_importance}.

When looking into the evaluation Table~\ref{tab:emb_importance} in Appendix~\ref{sec:appendix_emb_importance}, the trend is that, on average, weights higher than the initial 0.75 lead to worse results---more distance between the output image and the gold images or captions---, along all metrics except one, CLIP-T. 

But this average may conceal important differences among different examples.
For instance, in Figure~\ref{fig:examples_importance} the first row shows the results of the instruction ``Make the woman obese'' with increasing edit-distance strength.
The change conveyed by this instruction is suited to a gradual application.

In contrast, the alteration conveyed by the instruction in the second row, ``Let's add birds to the sky'', becomes effective only after reaching a threshold with a sufficiently high weight.
The change only occurs at 1.25, the highest weight experimented with.
Either there are birds in the sky or there are not, there is no way to almost have birds in the sky.

\subsection{Underlying LLMs}
\label{llms}

Table~\ref{tab:llms} presents the evaluation for the various LLMs studied. Each evaluation point took $\sim$12h on an NVIDIA A100 40GB GPU. Editing one image takes $\sim$30s.

Llama~2  has the worst performance both when comparison is made with images (CLIP-I) and with captions (CLIP-T).
This can be attributed to its being the only model considered that is not instruction-tuned and, therefore, has less capacity to follow instructions, as expected.

Gemma, Phi-2 and Mistral are close to each other in performance, with Phi-2 outperforming Gemma on three of the four metrics, and Mistral in one metric, despite it having less than half the parameters of Gemma and Mistral.

Mistral emerges as the best-performing model in this experiment, for all metrics except CLIP-I, where Phi-2 slightly outperforms it.
Mistral is also one BLEU point ahead of the second best models, when comparing output captions to gold test captions, which is larger than the differences among the other models.

This performance ranking of these models turns out to be as expected, as it is aligned with their ranking in the HuggingFace leaderboard.

Additionally, the first lines in Table~\ref{tab:llms} indicate scores from ablation studies. We experimented with using the embedding of the editing instruction only as the edit-direction embedding ("Instruction only" line), and with using the embedding of the after-edit caption only as the edit-direction embedding ("After-edit caption only" line), obtained with Mistral, is used. 

As expected, one obtains worse results in both cases than with the full system given important sources of information are missing.

The worst performance is found when directly applying the instruction to edit the image. This is as expected since no information about the input image is provided.

\begin{table*}[tp]
\centering
\begin{tabular}{lcccc}
\toprule
     & BLEU             & Cosine Sim.         & CLIP-T         & CLIP-I \\
\midrule
Instruction only &-&-&0.2818&0.8027\\
After-edit caption only &-&-&0.2897&0.8120\\
\midrule
Mistral &\textbf{10.2731}  &\textbf{0.7883 }&\textbf{0.2904 }& 0.8268                  \\
Phi-2& \phantom{1}9.2404& 0.7492         & 0.2880         & \textbf{0.8285}                  \\
Gemma  & \phantom{1}9.0551& 0.7351         & 0.2878         & 0.8215                  \\
Llama 2 & \phantom{1}8.5030& 0.7354         & 0.2819         & 0.8180                  \\
\bottomrule
\end{tabular}
\caption{Ablation study (top): using the instruction embedding as guide, using only the after-edit caption as guide; and experiments with different LLMs (bottom): using the full pipeline. Higher scores are better.}
\label{tab:llms}
\end{table*}



\subsection{Other systems}
\label{literature}

Table~\ref{tab:llms_literature} shows the scores of several state-of-the-art alternatives for this task.

Unlike ours, these approaches are supervised, requiring fine-tuning on some labelled dataset, which leads to models that follow the patterns found in the training dataset of MAGICBRUSH.

Note that, as mentioned above, MAGICBRUSH was created by mask-infilling.
Hence, a model trained on it will not learn to keep a close relationship with the input image.

This is particularly noticeable if the edit instruction requires modifying an object (e.g.~its colour, position, etc.), 
since mask-infilling hides the original object, and a brand new object will be generated in the gold target image. 

We refer back to the example in Figure~\ref{fig:dataset_example} where this can be observed, that is, the teddy bear in the ''Edited gold target´´ column (the target image from the MAGICBRUSH dataset) is severely different than the one in the input image.

This limitation negatively affects the reliability of the CLIP-I scores since we are comparing our output images to test target images that were created in such a manner.
Accordingly, CLIP-T appears to be a fairer metric since it is free from this drawback as it compares output images (embeddings) to test target captions (embeddings).

Under this metric, our best setup with Mistral has competitive performance with the state-of-the-art, supervised and competing approaches. 
With our 0.2904 score in CLIP-T being superior to the 0.2764 from InstructPix2Pix, the 0.2752 from HIVE, and the 0.2630 from EMU-Edit; and competitive with the 0.3040 from MGIE and the 0.3046 from ZONE.

Under CLIP-I, in turn, our system (0.8285 with Phi-2) is close in performance to both InstructPix2Pix (0.8524) and HIVE (0.8519), while being surpassed by EMU-Edit (0.8970), MGIE (0.9114), and ZONE (0.9269).

Still, this is an impressively good achievement since our method, being training-free, is being compared to supervised ones.

\begin{table*}[tp]
\centering
\begin{tabular}{lcccc}
\toprule
     & CLIP-T         & CLIP-I \\
   
\midrule
EMU-Edit  & 0.2630         & 0.8970 \\
HIVE & 0.2752         & 0.8519 \\
InstructPix2Pix & 0.2764         & 0.8524 \\
Ours (with Mistral) &0.2904 & 0.8268  \\
MGIE & 0.3040         & 0.9114 \\
ZONE & \textbf{0.3046}& \textbf{0.9269}\\
\bottomrule
\end{tabular}
\caption{Comparing our best approach (with Mistral) with the literature. Higher scores are better.}
\label{tab:llms_literature}
\end{table*}

\subsection{Manual evaluation}


To further assess the reliability of the automatic quantitative metrics for the evaluation results obtained, we ran two human qualitative evaluations, comparing our best setup (with Mistral) with InstructPix2Pix and with ZONE (the method with the highest CLIP-T and CLIP-I scores).\footnote{We did not find code available from MGIE to generate images for qualitative evaluation.}

Fifteen human annotators were provided with a questionnaire with 30 examples randomly selected from the MAGICBRUSH test set.

Each example included an input image, an edit request, and two output images, A and B, one generated by our method and the other by the other system (either InstructPi2Pix or ZONE), in a random order.

A characterization of the annotators and the questionnaire is in Appendix~\ref{sec:appendix_evaluation}, which also includes, in  Figure~\ref{fig:rand_examples}, some side-by-side examples of editing done with the three judged methods.

The annotators were asked to judge the result of the alteration requested and to indicate one of four possible answers: (i)~both are acceptable, (ii)~only A is acceptable, (iii)~only B is acceptable, and (iv)~none is acceptable. 

In the comparison with InstructPix2Pix, annotators found our approach acceptable in 33\% of the cases, against 24\% for InstructPix2Pix, thus reinforcing that our method performs better, confirming the relative ranking when using CLIP-T.

In turn, when compared against ZONE, this is deemed acceptable 37\% of the time, against 35\% for our approach, once again reinforcing the rankings obtained with CLIP-T.

It is also worth noting that ``none is acceptable'' was chosen 52\% of the time when comparing against InstructPix2Pix, and 44\% when comparing against ZONE, which indicates that there is still much room for improvement for the instruction-guided image editing task\footnote{An approach was considered acceptable if the annotator deemed either that only its result was acceptable or that both results were acceptable. 

In detail,the results obtained were: (i~InstructPix2Pix) both acceptable 9\%, only our approach acceptable 24\%, only InstructPix2Pix acceptable 15\%, and none acceptable 52\%; (ii~ZONE) both acceptable 16\%, only our approach acceptable 19\%, only ZONE acceptable 21\%, and none acceptable 44\%.}.

The evaluation results reported here and in previous sections demonstrate that not only our approach is an effective method for instruction-guided image editing, but it is also a competitive or superior alternative to recent approaches present in the literature, which set the previous state-of-the-art for this task but were supervised and much more resource-intensive.

\section{Conclusions}
\label{sec:conclusion}

This paper introduces a novel method for generic instruction-guided image editing that is based solely on inference over pre-existing models.  
Resorting to language models, the user-entered edit instruction is used to generate captions of images, and resorting to multimodal models, these captions are used to induce an edit-direction that supports the intended edit of the source image. 

This approach is free of task-specific training and of its respective resource-intensive overhead, and circumvents labelled data scarcity and biases introduced during supervised learning over them, which are drastic limitations to the previous methods in the literature.

The experiments reported here demonstrate that our method is effective for real-time image editing and that its performance is competitive with regards to alternative, state-of-the-art approaches recently proposed in the literature.

All its key components belong to very active research areas or topics that very likely will continue to get rapid advances in the short to medium term.
As their expected substantial improvements will induce substantial improvement for this approach, there is a great potential for its steady progress in the near future.

\section*{Acknowledgments}
PORTULAN CLARIN---Research Infrastructure for the Science and Technology of Language, funded by Lisboa 2020, Alentejo 2020 and FCT (PINFRA/22117/2016);
ACCELERAT.AI---Multilingual Intelligent Contact Centers, funded by IAPMEI (C625734525-00462629);
Language Driven Image Design with Diffusion, funded by FCT (2022.15880.CPCA.A1);
and IMPROMPT---Image Alteration with Language Prompts, funded by FCT (CPCA-IAC/AV/590897/2023).

\section{Limitations}
\label{sec:limitations}
This section discusses limitations, and potential remedies for them, concerning the results presented in this paper.

\subsection{Image reconstruction}
A first limitation concerns the DDIM inversion process. 
During this process some noise is introduced in the input image, and this may result in a reconstructed image that is slightly dissimilar from the original in aspects not related to the alteration requested.
This can be seen in some examples in Figure~\ref{fig:rand_examples}, and it is more prominent in the first and fourth examples in the left side, where the cows in the background were removed and the lighting fixtures disappeared, respectively. 
All other images may have some degree of dissimilarity, but the shape and position of the main objects in the image are eventually preserved.

Future improvements of research on this image inversion will expectedly allow for a boost in quality for image generation, and a~fortiori for an improved performance by our method.
Another possible improvement to consider consists in using null-text inversion~\cite{mokady:2023:nullinversion} in a future approach.

\subsection{Generation of captions}

Considering the results in Table~\ref{tab:llms}, BLEU and Cosine~Sim.\ scores increase along with the increase in quality of the output.
Given these metrics determine how close the after-edit caption is to the gold test caption, this indicates that improving the quality of these captions generated will improve the performance of the proposed method.

Improving the before-edit caption, in turn, will also deliver better results since, as it stands, it often only loosely describes the input image and sometimes even misses mentioning the object to be edited.
Having a better before-editing caption would also help the image inversion process.

Hence, better language generation models should have a positive impact in the performance of the proposed method.

Another step towards better captions is through the automatic finding of better prompts. 
Prompt optimization has gained traction as an effective mechanism for enhancing LLMs in several downstream tasks. 
Some exploratory work on this topic can be found in Appendix~\ref{sec:metaprompt}.


\subsection{Potential negative impact}

Instruction-guided image editing 
provides numerous opportunities for enhancing user engagement, creativity and accessibility. 

However, like other technologies, it may have a dual usage and the way it is used can significantly impact its effectiveness and its ethical standing.
Therefore, it is essential to use this technology in a responsible and ethical manner.

One of the major advantages of our approach to instruction-guided image editing is that it does not require any form of training, so no biases are introduced by our method.
However, it is important to note that the pre-existing pre-trained models that are eventually used by this method may have certain limitations and potential biases themselves that may affect its performance.
Therefore, users should be made aware of these limitations and be thus cautious to avoid any adverse effects.

\subsection{Further considerations}

Additional limitations can be read from what was presented in the paper, and are summarized here:  
(i)~The experiments are performed for one dataset only, partly due to the lack of further relevant datasets for the task studied in this paper. Other datasets are either too small, too domain-specific, or contain noise such as incorrect edits and low-quality images.
(ii)~Only the English language is taken into account, once again due to the test dataset available. However, it is worth noting that the proposed method itself is language independent, and not limited to English. Given that no training is required, any pre-trained model in any other language can be used.
(iii)~Only one test run was performed, due to hardware constraints since any testing run turned out to be quite time expensive given our experimental setup.

\subsection{Potential for progress}

All the key components of the method proposed here belong to very active research areas or topics that very likely will continue to get rapid advances in the short to medium term.
As their expected substantial improvements will induce substantial improvement for this approach, there is a great potential for its steady progress in the near future.

\bibliography{custom}

\begin{thebibliography}{49}
\providecommand{\natexlab}[1]{#1}

\bibitem[{Abdal et~al.(2020)Abdal, Qin, and Wonka}]{Abdal:2020:StyleGAN++}
Rameen Abdal, Yipeng Qin, and Peter Wonka. 2020.
\newblock {Image2StyleGAN++}: How to edit the embedded images?
\newblock \emph{arXiv preprint arXiv:1911.11544}.

\bibitem[{AI@Meta(2024)}]{llama3modelcard}
AI@Meta. 2024.
\newblock \href {https://github.com/meta-llama/llama3/blob/main/MODEL_CARD.md} {Llama 3 model card}.

\bibitem[{Bermano et~al.(2022)Bermano, Gal, Alaluf, Mokady, Nitzan, Tov, Patashnik, and Cohen-Or}]{Bermano:2022:StyleGANInversion}
Amit~H. Bermano, Rinon Gal, Yuval Alaluf, Ron Mokady, Yotam Nitzan, Omer Tov, Or~Patashnik, and Daniel Cohen-Or. 2022.
\newblock State-of-the-art in the architecture, methods and applications of {StyleGAN}.
\newblock \emph{arXiv preprint arXiv:2202.14020}.

\bibitem[{Brooks et~al.(2023)Brooks, Holynski, and Efros}]{brooks:2023:instructpix2pix}
Tim Brooks, Aleksander Holynski, and Alexei~A Efros. 2023.
\newblock {InstructPix2Pix}: Learning to follow image editing instructions.
\newblock In \emph{Proceedings of the IEEE/CVF Conference on Computer Vision and Pattern Recognition}, pages 18392--18402.

\bibitem[{Brown et~al.(2020)Brown, Mann, Ryder, Subbiah, Kaplan, Dhariwal, Neelakantan, Shyam, Sastry, Askell et~al.}]{Brown:2020:GPT3}
Tom~B Brown, Benjamin Mann, Nick Ryder, Melanie Subbiah, Jared Kaplan, Prafulla Dhariwal, Arvind Neelakantan, Pranav Shyam, Girish Sastry, Amanda Askell, et~al. 2020.
\newblock Language models are few-shot learners.
\newblock \emph{arXiv preprint arXiv:2005.14165}.

\bibitem[{Cheng et~al.(2020)Cheng, Gan, Li, Liu, and Gao}]{Cheng:2020:sequential}
Yu~Cheng, Zhe Gan, Yitong Li, Jingjing Liu, and Jianfeng Gao. 2020.
\newblock Sequential attention {GAN} for interactive image editing.
\newblock In \emph{Proceedings of the 28th ACM international conference on multimedia}, pages 4383--4391.

\bibitem[{Devlin et~al.(2018)Devlin, Chang, Lee, and Toutanova}]{Devlin:2018:Bert}
Jacob Devlin, Ming-Wei Chang, Kenton Lee, and Kristina Toutanova. 2018.
\newblock {BERT}: Pre-training of deep bidirectional transformers for language understanding.
\newblock \emph{arXiv preprint arXiv:1810.04805}.

\bibitem[{Dhariwal and Nichol(2021)}]{Dhariwal:2021:diffusion}
Prafulla Dhariwal and Alexander Nichol. 2021.
\newblock Diffusion models beat {GAN}s on image synthesis.
\newblock \emph{Advances in Neural Information Processing Systems}, 34.

\bibitem[{El-Nouby et~al.(2019)El-Nouby, Sharma, Schulz, Hjelm, Asri, Kahou, Bengio, and Taylor}]{El-nouby:2019:tell}
Alaaeldin El-Nouby, Shikhar Sharma, Hannes Schulz, Devon Hjelm, Layla~El Asri, Samira~Ebrahimi Kahou, Yoshua Bengio, and Graham~W Taylor. 2019.
\newblock Tell, draw, and repeat: Generating and modifying images based on continual linguistic instruction.
\newblock In \emph{Proceedings of the IEEE/CVF International Conference on Computer Vision}, pages 10304--10312.

\bibitem[{Fu et~al.(2023)Fu, Hu, Du, Wang, Yang, and Gan}]{fu:2023:MGIE}
Tsu-Jui Fu, Wenze Hu, Xianzhi Du, William~Yang Wang, Yinfei Yang, and Zhe Gan. 2023.
\newblock Guiding instruction-based image editing via multimodal large language models.
\newblock \emph{arXiv preprint arXiv:2309.17102}.

\bibitem[{Ge et~al.(2024)Ge, Zhao, Li, Ge, and Shan}]{ge:2024:seed-data}
Yuying Ge, Sijie Zhao, Chen Li, Yixiao Ge, and Ying Shan. 2024.
\newblock Seed-data-edit technical report: A hybrid dataset for instructional image editing.
\newblock \emph{arXiv preprint arXiv:2405.04007}.

\bibitem[{Goodfellow et~al.(2014)Goodfellow, Pouget-Abadie, Mirza, Xu, Warde-Farley, Ozair, Courville, and Bengio}]{Goodfellow:2014:GANs}
Ian Goodfellow, Jean Pouget-Abadie, Mehdi Mirza, Bing Xu, David Warde-Farley, Sherjil Ozair, Aaron Courville, and Yoshua Bengio. 2014.
\newblock Generative adversarial nets.
\newblock \emph{Advances in neural information processing systems}, 27.

\bibitem[{Gunasekar et~al.(2023)Gunasekar, Zhang, Aneja et~al.}]{gunasekar:2023:phi}
Suriya Gunasekar, Yi~Zhang, Jyoti Aneja, et~al. 2023.
\newblock Textbooks are all you need.
\newblock \emph{arXiv preprint arXiv:2306.11644}.

\bibitem[{Guo et~al.(2023)Guo, Wang, Guo, Li, Song, Tan, Liu, Bian, and Yang}]{guo2023connecting}
Qingyan Guo, Rui Wang, Junliang Guo, Bei Li, Kaitao Song, Xu~Tan, Guoqing Liu, Jiang Bian, and Yujiu Yang. 2023.
\newblock Connecting large language models with evolutionary algorithms yields powerful prompt optimizers.
\newblock \emph{arXiv preprint arXiv:2309.08532}.

\bibitem[{Hertz et~al.(2022)Hertz, Mokady, Tenenbaum et~al.}]{hertz:2022:prompt2prompt}
Amir Hertz, Ron Mokady, Jay Tenenbaum, et~al. 2022.
\newblock {Prompt-to-Prompt} image editing with cross attention control.
\newblock \emph{arXiv preprint arXiv:2208.01626}.

\bibitem[{Ho and Salimans(2022)}]{Ho:2022:ClassifierFree}
Jonathan Ho and Tim Salimans. 2022.
\newblock Classifier-free diffusion guidance.
\newblock \emph{arXiv preprint arXiv:2207.12598}.

\bibitem[{Hui et~al.(2024)Hui, Yang, Zhao, Shi, Wang, Wang, Zhou, and Xie}]{hui:2024:hq-data}
Mude Hui, Siwei Yang, Bingchen Zhao, Yichun Shi, Heng Wang, Peng Wang, Yuyin Zhou, and Cihang Xie. 2024.
\newblock Hq-edit: A high-quality dataset for instruction-based image editing.
\newblock \emph{arXiv preprint arXiv:2404.09990}.

\bibitem[{Jiang et~al.(2023)Jiang, Sablayrolles, Mensch et~al.}]{jiang:2023:mistral}
Albert~Q. Jiang, Alexandre Sablayrolles, Arthur Mensch, et~al. 2023.
\newblock Mistral {7B}.
\newblock \emph{arXiv preprint arXiv:2310.06825}.

\bibitem[{Jiang et~al.(2021)Jiang, Xu, Wang, Gao, Shi, Lin, and Liu}]{Jiang:2021:language}
Wentao Jiang, Ning Xu, Jiayun Wang, Chen Gao, Jing Shi, Zhe Lin, and Si~Liu. 2021.
\newblock Language-guided global image editing via cross-modal cyclic mechanism.
\newblock In \emph{Proceedings of the IEEE/CVF International Conference on Computer Vision}, pages 2115--2124.

\bibitem[{Kingma and Welling(2013)}]{Kingma:2013:VAE}
Diederik~P Kingma and Max Welling. 2013.
\newblock Auto-encoding variational {Bayes}.
\newblock \emph{arXiv preprint arXiv:1312.6114}.

\bibitem[{Kirillov et~al.(2023)Kirillov, Mintun, Ravi, Mao, Rolland, Gustafson, Xiao, Whitehead, Berg, Lo et~al.}]{kirillov:2023:sam}
Alexander Kirillov, Eric Mintun, Nikhila Ravi, Hanzi Mao, Chloe Rolland, Laura Gustafson, Tete Xiao, Spencer Whitehead, Alexander~C Berg, Wan-Yen Lo, et~al. 2023.
\newblock Segment anything.
\newblock In \emph{Proceedings of the IEEE/CVF International Conference on Computer Vision}, pages 4015--4026.

\bibitem[{Kong et~al.(2024)Kong, Amba~Hombaiah, Zhang, Mei, and Bendersky}]{kong2024prewrite}
Weize Kong, Spurthi Amba~Hombaiah, Mingyang Zhang, Qiaozhu Mei, and Michael Bendersky. 2024.
\newblock Prewrite: Prompt rewriting with reinforcement learning.
\newblock \emph{arXiv e-prints}, pages arXiv--2401.

\bibitem[{Lester et~al.(2021)Lester, Al-Rfou, and Constant}]{lester2021power}
Brian Lester, Rami Al-Rfou, and Noah Constant. 2021.
\newblock The power of scale for parameter-efficient prompt tuning.
\newblock In \emph{Proceedings of the 2021 Conference on Empirical Methods in Natural Language Processing}, pages 3045--3059.

\bibitem[{Li et~al.(2022)Li, Li, Xiong, and Hoi}]{li:2022:blip}
Junnan Li, Dongxu Li, Caiming Xiong, and Steven Hoi. 2022.
\newblock {BLIP}: Bootstrapping language-image pre-training for unified vision-language understanding and generation.
\newblock In \emph{International Conference on Machine Learning}, pages 12888--12900. PMLR.

\bibitem[{Li et~al.(2024)Li, Zeng, Feng, Gao, Liu, Liu, Li, Tang, Hu, Liu et~al.}]{li:2024:zone}
Shanglin Li, Bohan Zeng, Yutang Feng, Sicheng Gao, Xiuhui Liu, Jiaming Liu, Lin Li, Xu~Tang, Yao Hu, Jianzhuang Liu, et~al. 2024.
\newblock Zone: Zero-shot instruction-guided local editing.
\newblock In \emph{Proceedings of the IEEE/CVF Conference on Computer Vision and Pattern Recognition}, pages 6254--6263.

\bibitem[{Lin et~al.(2014)Lin, Maire, Belongie et~al.}]{lin:2014:coco}
Tsung-Yi Lin, Michael Maire, Serge Belongie, et~al. 2014.
\newblock Microsoft {COCO}: Common objects in context.
\newblock In \emph{Proceedings of ECCV 2014: 13th European Conference on Computer Vision}, pages 740--755. Springer.

\bibitem[{Mesnard et~al.(2024)Mesnard, Hardin, Dadashi, Bhupatiraju, Pathak, Sifre, Rivi{\`e}re, Kale, Love et~al.}]{Mesnard:2024:Gemma}
Thomas Mesnard, Cassidy Hardin, Robert Dadashi, Surya Bhupatiraju, Shreya Pathak, Laurent Sifre, Morgane Rivi{\`e}re, Mihir~Sanjay Kale, Juliette Love, et~al. 2024.
\newblock Gemma: Open models based on gemini research and technology.
\newblock \emph{arXiv preprint arXiv:2403.08295}.

\bibitem[{Mokady et~al.(2023)Mokady, Hertz, Aberman et~al.}]{mokady:2023:nullinversion}
Ron Mokady, Amir Hertz, Kfir Aberman, et~al. 2023.
\newblock Null-text inversion for editing real images using guided diffusion models.
\newblock In \emph{Proceedings of the IEEE/CVF Conference on Computer Vision and Pattern Recognition}, pages 6038--6047.

\bibitem[{Papineni et~al.(2002)Papineni, Roukos, Ward, and Zhu}]{Papineni:2002:BLEU}
Kishore Papineni, Salim Roukos, Todd Ward, and Wei-Jing Zhu. 2002.
\newblock {BLEU}: a method for automatic evaluation of machine translation.
\newblock In \emph{Proceedings of the 40th annual meeting of the Association for Computational Linguistics}, pages 311--318.

\bibitem[{Parmar et~al.(2023)Parmar, Kumar~Singh, Zhang et~al.}]{parmar:2023:pix2pix-zero}
Gaurav Parmar, Krishna Kumar~Singh, Richard Zhang, et~al. 2023.
\newblock Zero-shot image-to-image translation.
\newblock In \emph{ACM SIGGRAPH 2023 Conference Proceedings}, pages 1--11.

\bibitem[{Prasad et~al.(2023)Prasad, Hase, Zhou, and Bansal}]{prasad2023grips}
Archiki Prasad, Peter Hase, Xiang Zhou, and Mohit Bansal. 2023.
\newblock Grips: Gradient-free, edit-based instruction search for prompting large language models.
\newblock In \emph{Proceedings of the 17th Conference of the European Chapter of the Association for Computational Linguistics}, pages 3845--3864.

\bibitem[{Ramesh et~al.(2022)Ramesh, Dhariwal, Nichol, Chu, and Chen}]{Ramesh:2022:DALLE2}
Aditya Ramesh, Prafulla Dhariwal, Alex Nichol, Casey Chu, and Mark Chen. 2022.
\newblock Hierarchical text-conditional image generation with clip latents.
\newblock \emph{arXiv preprint arXiv:2204.06125}.

\bibitem[{Rombach et~al.(2022)Rombach, Blattmann, Lorenz et~al.}]{rombach:2022:stable_diffusion}
Robin Rombach, Andreas Blattmann, Dominik Lorenz, et~al. 2022.
\newblock High-resolution image synthesis with latent diffusion models.
\newblock In \emph{Proceedings of the IEEE/CVF Conference on Computer Vision and Pattern Recognition}, pages 10684--10695.

\bibitem[{Santos et~al.(2022{\natexlab{a}})Santos, Branco, and Silva}]{santos:2022:lxdrim}
Rodrigo Santos, Ant{\'o}nio Branco, and Jo{\~a}o Silva. 2022{\natexlab{a}}.
\newblock Cost-effective language driven image editing with lx-drim.
\newblock In \emph{Proceedings of the First Workshop on Performance and Interpretability Evaluations of Multimodal, Multipurpose, Massive-Scale Models}, pages 31--43.

\bibitem[{Santos et~al.(2022{\natexlab{b}})Santos, Branco, and Silva}]{santos:2022:languageDriven}
Rodrigo Santos, Ant{\'o}nio Branco, and Jo{\~a}o Silva. 2022{\natexlab{b}}.
\newblock Language driven image editing via transformers.
\newblock In \emph{2022 IEEE 34th International Conference on Tools with Artificial Intelligence (ICTAI)}, pages 909--914. IEEE.

\bibitem[{Santos et~al.(2024)Santos, Silva, and Branco}]{santos:2024:leveragingLLMs}
Rodrigo Santos, Jo{\~a}o Silva, and Ant{\'o}nio Branco. 2024.
\newblock Leveraging llms for on-the-fly instruction guided image editing.
\newblock In \emph{Progress in Artificial Intelligence (EPIA)}, pages 28--40.

\bibitem[{Sheynin et~al.(2024)Sheynin, Polyak, Singer, Kirstain, Zohar, Ashual, Parikh, and Taigman}]{sheynin:2024:emu}
Shelly Sheynin, Adam Polyak, Uriel Singer, Yuval Kirstain, Amit Zohar, Oron Ashual, Devi Parikh, and Yaniv Taigman. 2024.
\newblock Emu edit: Precise image editing via recognition and generation tasks.
\newblock In \emph{Proceedings of the IEEE/CVF Conference on Computer Vision and Pattern Recognition}, pages 8871--8879.

\bibitem[{Song et~al.(2020)Song, Meng, and Ermon}]{song:2020:DDIM}
Jiaming Song, Chenlin Meng, and Stefano Ermon. 2020.
\newblock Denoising diffusion implicit models.
\newblock \emph{arXiv preprint arXiv:2010.02502}.

\bibitem[{Srivastava et~al.(2023)Srivastava, Rastogi et~al.}]{srivastava2023beyond}
Aarohi Srivastava, Abhinav Rastogi, et~al. 2023.
\newblock Beyond the imitation game: Quantifying and extrapolating the capabilities of language models.
\newblock \emph{Transactions on Machine Learning Research}.

\bibitem[{Touvron et~al.(2023)Touvron, Martin, Stone et~al.}]{touvron:2023:llama2}
Hugo Touvron, Louis Martin, Kevin Stone, et~al. 2023.
\newblock {LLaMA~2}: Open foundation and fine-tuned chat models.
\newblock \emph{arXiv preprint arXiv:2307.09288}.

\bibitem[{Vaswani et~al.(2017)Vaswani, Shazeer, Parmar, Uszkoreit, Jones, Gomez, Kaiser, and Polosukhin}]{Vaswani:2017:Transformer}
Ashish Vaswani, Noam Shazeer, Niki Parmar, Jakob Uszkoreit, Llion Jones, Aidan~N Gomez, {\L}ukasz Kaiser, and Illia Polosukhin. 2017.
\newblock Attention is all you need.
\newblock In \emph{Advances in neural information processing systems}, pages 5998--6008.

\bibitem[{Wasserman et~al.(2024)Wasserman, Rotstein, Ganz, and Kimmel}]{wasserman:2024:inpaint}
Navve Wasserman, Noam Rotstein, Roy Ganz, and Ron Kimmel. 2024.
\newblock Paint by inpaint: Learning to add image objects by removing them first.
\newblock \emph{arXiv preprint arXiv:2404.18212}.

\bibitem[{Wei et~al.(2022)Wei, Wang, Schuurmans, Bosma, Xia, Chi, Le, Zhou et~al.}]{wei2022chain}
Jason Wei, Xuezhi Wang, Dale Schuurmans, Maarten Bosma, Fei Xia, Ed~Chi, Quoc~V Le, Denny Zhou, et~al. 2022.
\newblock Chain-of-thought prompting elicits reasoning in large language models.
\newblock \emph{Advances in neural information processing systems}, 35:24824--24837.

\bibitem[{Xia et~al.(2022)Xia, Zhang, Yang, Xue, Zhou, and Yang}]{Xia:2022:GANInversion}
Weihao Xia, Yulun Zhang, Yujiu Yang, Jing-Hao Xue, Bolei Zhou, and Ming-Hsuan Yang. 2022.
\newblock {GAN} inversion: A survey.
\newblock \emph{arXiv preprint arXiv:2101.05278}.

\bibitem[{Yang et~al.(2023)Yang, Wang, Lu, Liu, Le, Zhou, and Chen}]{yang2023large}
Chengrun Yang, Xuezhi Wang, Yifeng Lu, Hanxiao Liu, Quoc~V Le, Denny Zhou, and Xinyun Chen. 2023.
\newblock Large language models as optimizers.
\newblock \emph{arXiv preprint arXiv:2309.03409}.

\bibitem[{Yao et~al.(2024)Yao, Yu, Zhao, Shafran, Griffiths, Cao, and Narasimhan}]{yao2024tree}
Shunyu Yao, Dian Yu, Jeffrey Zhao, Izhak Shafran, Tom Griffiths, Yuan Cao, and Karthik Narasimhan. 2024.
\newblock Tree of thoughts: Deliberate problem solving with large language models.
\newblock \emph{Advances in Neural Information Processing Systems}, 36.

\bibitem[{Zhang et~al.(2024)Zhang, Mo, Chen et~al.}]{zhang:2024:magicbrush}
Kai Zhang, Lingbo Mo, Wenhu Chen, et~al. 2024.
\newblock {MAGICBRUSH}: A manually annotated dataset for instruction-guided image editing.
\newblock \emph{Advances in Neural Information Processing Systems}, 36.

\bibitem[{Zhang et~al.(2023)Zhang, Yang, Feng et~al.}]{zhang:2023:hive}
Shu Zhang, Xinyi Yang, Yihao Feng, et~al. 2023.
\newblock {HIVE}: Harnessing human feedback for instructional visual editing.
\newblock \emph{arXiv preprint arXiv:2303.09618}.

\bibitem[{Zhou et~al.(2022)Zhou, Muresanu, Han, Paster, Pitis, Chan, and Ba}]{zhou2022large}
Yongchao Zhou, Andrei~Ioan Muresanu, Ziwen Han, Keiran Paster, Silviu Pitis, Harris Chan, and Jimmy Ba. 2022.
\newblock Large language models are human-level prompt engineers.
\newblock In \emph{Proceedings of The Eleventh International Conference on Learning Representations}.

\end{thebibliography}


\onecolumn
\appendix

\section{Image generation models}
\label{sec:appendix_sd}
This section details the two diffusion image generation models used in this work, namely Stable Diffusion 1.4 and 1.5.

\paragraph{Stable Diffusion 1.4}
The Stable Diffusion model is a text-conditioned image generator model that combines an autoencoder with a diffusion model to create a latent diffusion model.
The autoencoder encodes images into latent representations with a reduced dimensionality when compared to the input image, reducing the computational needs during the training phase.
Text prompts, on the other hand, are encoded using a text encoder and are then cross-attended by the UNet backbone of the latent diffusion model. 
Finally, the loss is computed using a reconstruction objective between the noise added to the latent representation and the prediction made by the UNet.

Stable Diffusion 1.4 (\url{https://huggingface.co/CompVis/stable-diffusion-v1-4}) had several rounds of training on the LAION dataset (\url{https://laion.ai/}), with each round changing the input image dimension, aesthetic score, and the probability of dropping the text-conditioning to improve classifier-free guidance. 

\paragraph{Stable Diffusion 1.5}
SD 1.5, in turn, has the same architecture and even the same starting point as 1.4, with the difference being how long the model was fine-tuned on top of SD 1.2.
The 1.4 version is fine-tuned for 225 thousand steps at resolution 512x512 on ``laion-aesthetics v2 5+'' with a 10\% probability of dropping the text-conditioning, and version 1.5 for 595 thousand steps.

As demonstrated in Section~\ref{sec:appendix_emb_importance} Stable Diffusion 1.4 has better performance than 1.5 in our approach, therefore, we will adopt SD 1.4 for most of the experiments in this paper.

\section{Large language models}
\label{sec:appendix_llms}

Here we give additional details on the large language models that we used in our experiments.

\paragraph{Gemma} \cite{Mesnard:2024:Gemma}, trained on a diverse 6~Trillion token dataset comprising web documents, code and mathematical texts.
We resorted to the 7~Billion parameter instruction-tuned decoder-only model, named \emph{gemma-7b-it} (\url{https://huggingface.co/google/gemma-7b-it}).
This model uses a chat template, which we employ during inference.

\paragraph{Llama~2} \cite{touvron:2023:llama2}, of which we used the 7~Billion parameter, pre-trained-only model, \emph{Llama-2-7b} (\url{https://huggingface.co/meta-llama/Llama-2-7b-hf}).
This model was trained with a mix of publicly available data totalling 2~Trillion tokens.
While its chat versions employ supervised fine-tuning and reinforcement learning with human feedback for alignment with human preferences in helpfulness and safety, the pre-trained-only model does not.
This results in a less constrained model, but it may also cause it to disperse from the task at hand. 
Since this model is a pre-trained-only no chat template is needed.

\paragraph{Mistral} \cite{jiang:2023:mistral} fine-tuned on various HuggingFace instruction datasets.
We resorted to the 7~Billion \emph{Mistral-7B-Instruct-v0.2} model (\url{https://huggingface.co/mistralai/Mistral-7B-Instruct-v0.2}) and used the respective chat template during inference.

\paragraph{Phi-2} \cite{gunasekar:2023:phi} is a compact 2.7~Billion model (\url{https://huggingface.co/microsoft/phi-2}).
Despite its size, it offers a competitive performance with respect to models several times its size.
It was trained on 250~Billion tokens, obtained through a combination of NLP synthetic data created by GPT-3.5 and filtered web data from Falcon RefinedWeb and SlimPajama, which was assessed by GPT-4.
This model was not fine-tuned through reinforcement learning from human feedback and does not have guardrails.

\subsubsection*{Model ranking}

A ranking of these models in terms of their performance can be found in the HuggingFace leaderboard (\url{https://huggingface.co/spaces/HuggingFaceH4/open_llm_leaderboard}) which assesses several LLMs that are trained under the same criteria and tested on the same benchmarks, including reasoning challenges and adversarial tasks.
As of this writing, the models above rank on the leaderboard with the following overall scores: \emph{Llama-2-7b} with 50.97, \emph{Gemma-7b-it} with 53.56, \emph{Phi-2} with 61.33, and \emph{Mistral-7B-Instruct-v0.2} with 65.71.

\begin{table}[tp]
\centering
\begin{tabular}{l @{\hspace{5ex}} cc @{\hspace{5ex}} cc}
\toprule
                           & \multicolumn{2}{l}{\hspace{3.6ex}CLIP-T}       & \multicolumn{2}{l}{\hspace{3.1ex}CLIP-I} \\
                           & avg              & stdev     & avg             & stdev \\
\midrule

\addlinespace
\multicolumn{5}{l}{Our model, 0-shot}                                                \\ 
\midrule
1-caption                  & 0.2751           & 0.0007    & 0.8021          & 0.0013 \\
1-caption; w. BLIP caption & 0.2795           & 0.0003    & 0.8255          & 0.0044 \\
2-caption; w. BLIP caption & 0.2796           & 0.0001    & 0.8329          & 0.0003 \\
4-caption; w. BLIP caption & 0.2799           & 0.0003    & 0.8347          & 0.0007 \\ 
\midrule

\addlinespace
\multicolumn{5}{l}{Our model, 1-shot}                                                \\ \midrule
1-caption                  & 0.2772           & 0.0012    & 0.8093          & 0.0023 \\
1-caption; w. BLIP caption & \textbf{0.2817}  & 0.0003    & 0.8310          & 0.0002 \\
2-caption; w. BLIP caption & 0.2800           & 0.0008    & 0.8328          & 0.0016 \\
4-caption; w. BLIP caption & 0.2797           & 0.0002    & 0.8348          & 0.0009 \\
\midrule

\addlinespace
\multicolumn{5}{l}{Our model, 3-shot}                                                \\
\midrule
1-caption                  & 0.2762           & 0.0003    & 0.8119          & 0.0032 \\
1-caption; w. BLIP caption & 0.2798           & 0.0001    & 0.8251          & 0.0001 \\
2-caption; w. BLIP caption & 0.2797           & 0.0000    & 0.8348          & 0.0010 \\
4-caption; w. BLIP caption & 0.2790           & 0.0000    & \textbf{0.8350} & 0.0011 \\
\bottomrule
\end{tabular}
\caption{CLIP cosine distance scores, averaged over two prompts, for different number of few-shot examples and different number of captions. The best scores are shown in bold.}
\label{tab:n_captions_results}
\end{table}

\section{Number of captions}
\label{sec:appendix_n_captions}
Table~\ref{tab:n_captions_results} presents the results obtained during the experiments on the number of captions to be generated. 
Two runs were performed, using Phi-2 as the LLM, with distinct prompts: 
\begin{itemize}
    \item A terse prompt:
    ``Given the transformation `[TRANSFORMATION]' generate [NUMBER] image captions for before and after the transformation.'';
    \item and a more expressive and detailed prompt:
    ``Employing the specified method `[TRANSFORMATION]', craft [NUMBER] pairs of descriptive captions delineating the images both prior to and following the application of the transformation process, elucidating the changes brought about.''.
\end{itemize}

Figure~\ref{fig:prompt_example} showcases examples of the output from the LLM.

\begin{figure}[tp]
\begin{Verbatim}[frame=single,fontsize=\small]
Instruct: Given the transformation `Make the cat a dog', generate 2 image captions for before and 
after the transformation.
Output: Before transformation

Caption 1: A photo of a tabby cat sleeping.
Caption 2: A cat playing with a ball of yarn.

After transformation

Caption 1: A photo of a cute dog.
Caption 2: A dog chewing on a bone.
\end{Verbatim}
\vspace{-3ex}
    \caption{Example of the output generated by the language model.}
    \label{fig:prompt_example}
\end{figure}

The values in the ``avg''  columns are the average between scores obtain by the two studied prompts, and the ``stdev'' columns their standard deviation.

As an overall trend, when evaluating against the gold image through CLIP-I, the more examples in the few-shot prompt the better; in contrast, with CLIP-T, the better performance is obtained with only one-shot.

Similarly, with CLIP-I, the larger the number of captions generated the better the performance; in contrast, with CLIP-T, the better performance is obtained with only one caption generated (generated by BLIP).


\section{Edit-embedding weight}
\label{sec:appendix_emb_importance}
Table~\ref{tab:emb_importance} presents the results obtained in the experiments on the impact of the edit-embedding weight, as well as the performance between different Stable Diffusion models, where ``Tgt'' means that the comparison is with the target gold image/caption, and ``Src'' with the input image/caption.

The CLIP-T Tgt metric shows a trend opposite to the other three metrics when edit-distance is increased.
The more changes are done to the input image, the further away the output image will be from it, but it is also possible for the changes to lead to a result that is more distant from the gold test image.
Therefore, the presence of more alterations is positively accommodated by CLIP-T Tgt---which compares output image and gold test \emph{caption}---and can be negatively accommodated by CLIP-I Tgt---which, in turn, compares output image and test \emph{image}.

This, once again, raises the concern that CLIP-I may not be the best evaluation metric since the requested change can be performed in almost infinite ways.
For instance, there will be changes in CLIP-I score with as little, and as irrelevant, alterations as a slight shift of the place where the birds are added to the sky in Figure~\ref{fig:examples_importance}. 

CLIP-T Tgt, in turn, compares to a ground language representation of the target, rather than a visual one.
It seems thus to be a more robustly abstract and general representation of the intended target than a (perhaps too specific) image fixed to serve as the ground test.

Finally, the experiments with SD 1.5 showcase the same trends as above, despite its slightly worse performance when compared with SD 1.4. For more details see \cite{santos:2024:leveragingLLMs}.
\begin{table}[tp]
\centering
\begin{tabular}{lcccc}
\toprule
$w$ & CLIP-I Tgt & CLIP-T Tgt & CLIP-I Src & CLIP-T Src \\
\midrule
0.75 & \textbf{0.8374}    & 0.2804   & \textbf{0.8555}    & \textbf{0.2874}   \\
1    & 0.8366    & 0.2841   & 0.8501    & 0.2863   \\
1.25 & 0.8285    & \textbf{0.2880}   & 0.8340    & 0.2822   \\
\midrule
0.75 & 0.8335    & 0.2796   & 0.8510    & 0.2865   \\
1    & 0.8356    & 0.2847   & 0.8480    & 0.2856   \\
1.25 & 0.8196    & 0.2870   & 0.8240    & 0.2805  \\
\bottomrule
\end{tabular}
\caption{Comparison SD 1.4 (top) vs.\ 1.5 (bottom), along three edit-distance weights ($w$), according to four distance metrics (larger is better), under Phi-2.}
\label{tab:emb_importance}
\end{table}

\section{Human evaluation details}
\label{sec:appendix_evaluation}

Human annotators were all volunteers. They were properly informed of the purpose of the collected data and all consented to the use of their responses for evaluating the outcome of research on an AI tool and to its scientific publication. 

The vast majority of the annotators had a higher education degree. They were \textasciitilde 62\% female and \textasciitilde 38\% male.

The exact written instruction provided to the annotators is as follows:
\begin{center}
\small
\begin{tabular}{|p{0.01cm}p{0.81\columnwidth}p{0.01cm}|}
\hline
&&\\
&Please judge how well the editing request was performed between system A and system B. Only keep the most appropriate answer, i.e.:& \\
&“A is best” if A satisfies the request better than B;&\\
&“B is best” if B satisfies the request better than A;&\\
&“Both are ok” if both equally satisfy the request;&\\
&“None are ok" if none satisfies the request.& \\
&&\\

\hline
\end{tabular}
\end{center}

\begin{figure*}[ht]
    \centering
    \includegraphics[width=1\linewidth]{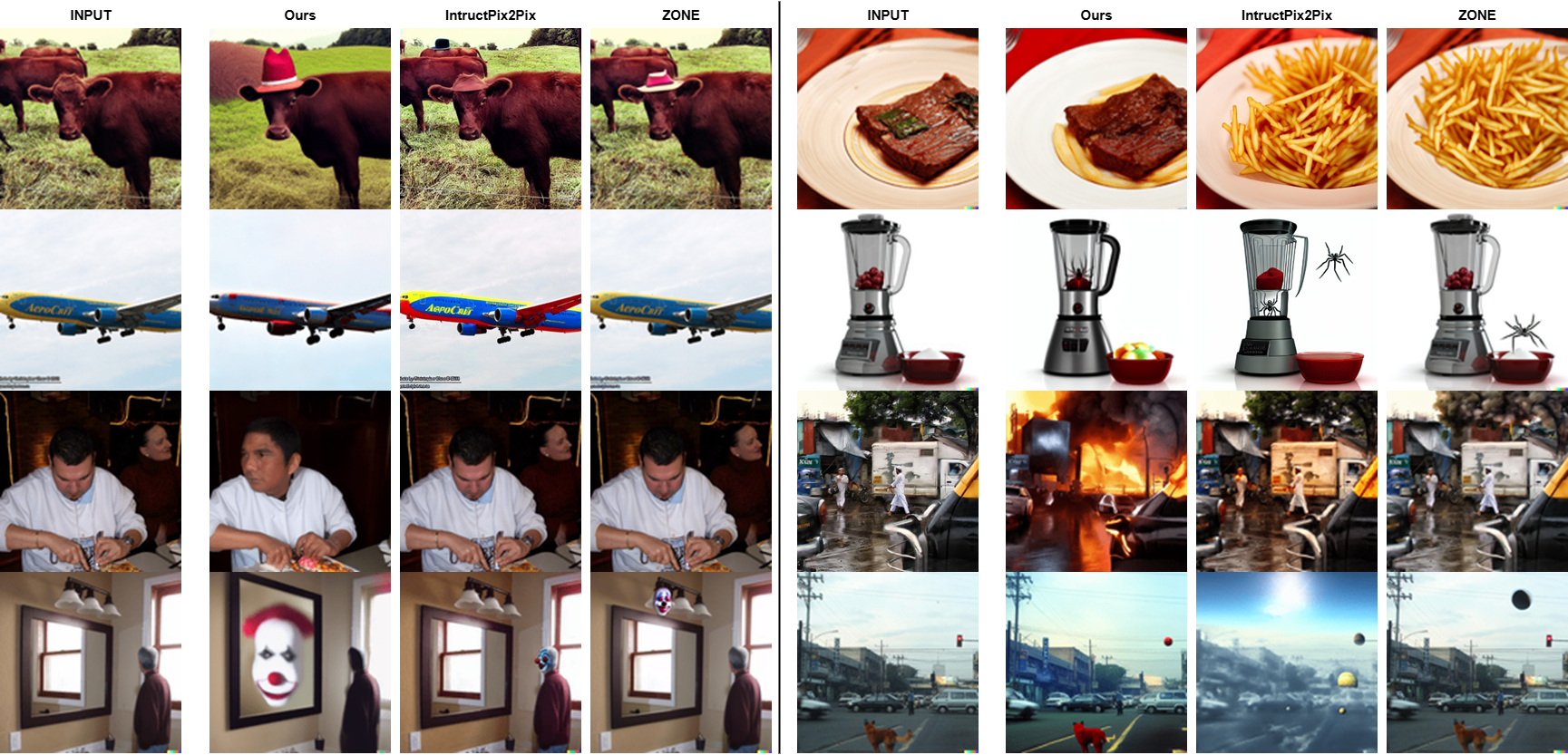}
    \caption{Examples from MAGICBRUSH test set (first columns) edited with our method (second columns), InstructPix2Pix (third columns), and ZONE (fourth columns). Edit-requests left side: ``Have the cow wear a hat.''; ``Change the blue and yellow to red and white plane.'';  ``Make the man look to the camera.'';``Put a clown face on the mirror.''. Edit-requests right-side: ``It should have french fries on the plate.''; ``Add a spider next to the blender.''; ``Add fire to the buildings.''; ``Put an exotic planet in the sky.''. These images were obtained through 100 DDIM inversion steps, 100 DDIM image generation steps and with captions generated with Mistral.}
    \label{fig:rand_examples}
\end{figure*}

Figure \ref{fig:rand_examples} shows some random examples taken from the MAGICBRUSH test set, used for manual evaluation.

\section{Meta-prompt}
\label{sec:metaprompt}

Our method resorts to language and image models only for inference.
While fully unsupervised, it still relies on a prompt created by a human to generate the target captions. 
A further step towards unwavering unsupervision, consists in  dispensing not only with supervised learning over labelled datasets, but also with human intervention to design that prompt.

In a second set of experiments, presented in this Section, our method resorts to prompts that are automatically created, being extended with a prompt optimization technique for that purpose.

\subsection{Related work}

Prompt optimization has gained traction as an effective mechanism for enhancing LLMs in several downstream tasks \cite{lester2021power,srivastava2023beyond}.
Recent studies have introduced techniques such as chain-of-thought \cite{wei2022chain} and tree-of-thoughts \cite{yao2024tree}, searching through a pool of prompt candidates generated by an LLM \cite{zhou2022large}, applying iterative local edit operations at a syntactic phrase-level split within the prompts \cite{prasad2023grips}, employing reinforcement learning to rewrite prompts \cite{kong2024prewrite} or evolutionary operators over a prompt population for optimization \cite{guo2023connecting}.

In particular, \citet{yang2023large} introduced the state-of-the-art OPRO technique, leveraging LLMs as optimizers through meta-prompts. 
We extend our image editing method with prompts that are automatically generated on the basis of such met-prompting.

\subsection{Meta-prompting}

\begin{figure*}[ht]
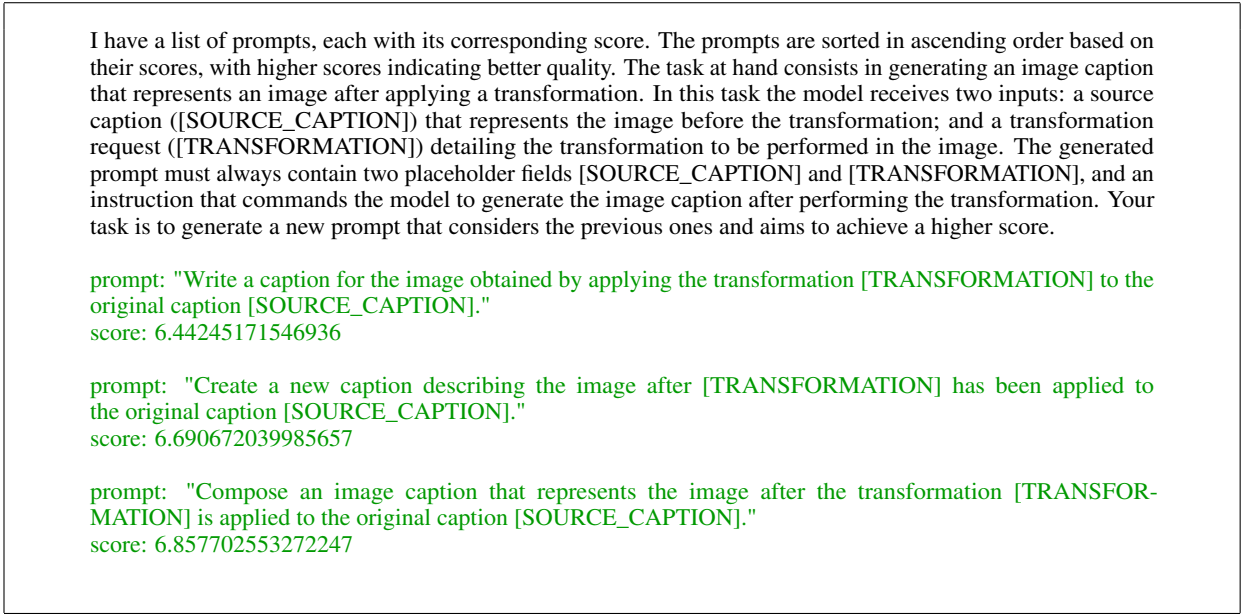

\small
\begin{tabular}{|p{0.5cm}p{14cm}p{0.5cm}|}
\hline

& \multicolumn{1}{c}{} &  \\
& 

I have a list of prompts, each with its corresponding score. The prompts are sorted in ascending order based on their scores, with higher scores indicating better quality. The task at hand consists in generating an image caption that represents an image after applying a transformation. In this task the model receives two inputs: a source caption ([SOURCE\_CAPTION]) that represents the image before the transformation; and a transformation request ([TRANSFORMATION]) detailing the transformation to be performed in the image. The generated prompt must always contain two placeholder fields [SOURCE\_CAPTION] and [TRANSFORMATION], and an instruction that commands the model to generate the image caption after performing the transformation. Your task is to generate a new prompt that considers the previous ones and aims to achieve a higher score. 
\newline
{\color[HTML]{009901} 

prompt: "Write a caption for the image obtained by applying the transformation [TRANSFORMATION] to the original caption [SOURCE\_CAPTION]." 
\newline
score: 6.44245171546936
\newline
\newline
prompt: "Create a new caption describing the image after [TRANSFORMATION] has been applied to the original caption [SOURCE\_CAPTION]."\newline
score: 6.690672039985657
\newline
\newline
prompt: "Compose an image caption that represents the image after the transformation [TRANSFORMATION] is applied to the original caption [SOURCE\_CAPTION]."\newline
score: 6.857702553272247
\newline
\newline
}
& \\
 \hline
\end{tabular}
\newline
\caption{Meta-prompt: in black, the top paragraph with the meta-instruction used in the experiments; below, in green, the list of top performing prompts after the last optimization step, and the respective scores. The best scoring prompt, at the bottom, is the final prompt that was automatically obtained and used for the results presented in Table~\ref{tab:meta_prompt_results}. Experimentation took \textasciitilde 2 days on three NVIDIA A100 40GB GPUs.}
\label{tab:meta_prompt_example}
\end{figure*}



\begin{algorithm*}
\small
\begin{algorithmic}[1]
\State \textbf{Input:} Dataset $D$, each examples contains a before-edit caption $bc$ and the editing request $er$; meta-prompt $metaP$ with the description of the optimization task.
\State \textbf{Output:} $history$: list of best scored prompts

\State $history \gets \emptyset$
\For{step in number of optimization steps} \Comment{We use 20 optimizations steps}
    \State $input \gets metaP \oplus history$ \Comment{$\oplus$ : concatenation}
    \State $newP \gets LLM(input)$  \Comment{We generate 2 prompts; LLM() invokes the llm to generate a continuation} 
    \State $allP \gets newP \oplus history$
    \State $E \gets$ random subset from $D$ \Comment{We select 8 examples}
    \For{each prompt $p_i$ in $allP$}
        \For{each example $e_j$ in $E$}
            \State $transP \gets template(p_i, bc, er)$ \Comment{template() fills the placeholders of the generated prompts}  
            \State $ac \gets LLM(transP)$ \Comment{$ac$ is the after-edit caption}
            \State $score_{ij} \gets evaluate(e_j,ac)$ \Comment{evaluate() edits the image and evaluates the result}
        \EndFor
        \State  $score_i \gets \sum_j score_{ij}$
    \EndFor
    \State $history \gets$ top best evaluated prompts \Comment{We keep the top-3 best prompts}
\EndFor
\end{algorithmic}
\caption{Meta-prompt algorithm}
\label{alg:meta-prompt}
\end{algorithm*}

Meta-prompting is a method for optimizing prompts using LLMs and natural language descriptions to guide it. 
Through iterative refinement over a meta-prompt describing a task to be optimized, an LLM generates new prompts based on a history of previous prompts and their scores.
This process can automatically discover high-performance prompts by exploring the optimization trajectory and searching for better prompts.  

The integration of the meta-prompting technique in our image editing method aims to enhance editing performance by refining the prompt eventually used to generate the after-edit caption.
This is achieved by prompting the LLM to generate a target caption based on the source caption and the editing instruction and by iteratively generating and evaluating other possible prompts for the same purpose, using the instructions delivered so far and their performance scores to guide the eventual finding of the optimal prompt for this concrete task.

An example of a meta-prompt is provided in Figure~\ref{tab:meta_prompt_example}, and a detailed description of this optimization via meta-prompting in Algorithm~\ref{alg:meta-prompt}. 

\subsection{A larger language model}

Meta-prompting requires a significantly larger model than the ones used in the first set of experiments. 
In this second set of experiments, we resort to a 70 Billion parameter model, Llama-3-70B-Instruct (\url{https://huggingface.co/meta-llama/Meta-Llama-3-70B-Instruct}) \cite{llama3modelcard}, using 4-bit quantization.
This model was pre-trained over 15 trillion tokens from publicly available sources followed by supervised fine-tuning and reinforcement learning with human feedback to align it with human preferences.
We opted for this model after preliminary experiments with smaller models had failed to deliver any sensible meta-prompting optimization.

We experimented with a maximum of 20 steps for optimization. At each step, two new prompts are generated and evaluation is undertaken.
Each of these 2 prompts is evaluated on 8 examples randomly sampled from the MAGICBRUSH development set, which are used to (re-)evaluate also the 3 prompts retained in the prompt history, which had been the best so far.
These 5 prompts (2 new + 3 previous) are re-ranked according to this new evaluation and only the 3 best are retained at each iteration step.

To guide the optimization process through these steps, CLIP-I was used as the evaluation metric since the MAGICBRUSH development set does not have captions for the before and after-edit images and thus does support any of the other metrics, which rely totally or partly on captions.




\begin{figure}
\centering
    \includegraphics[width=0.5\textwidth]{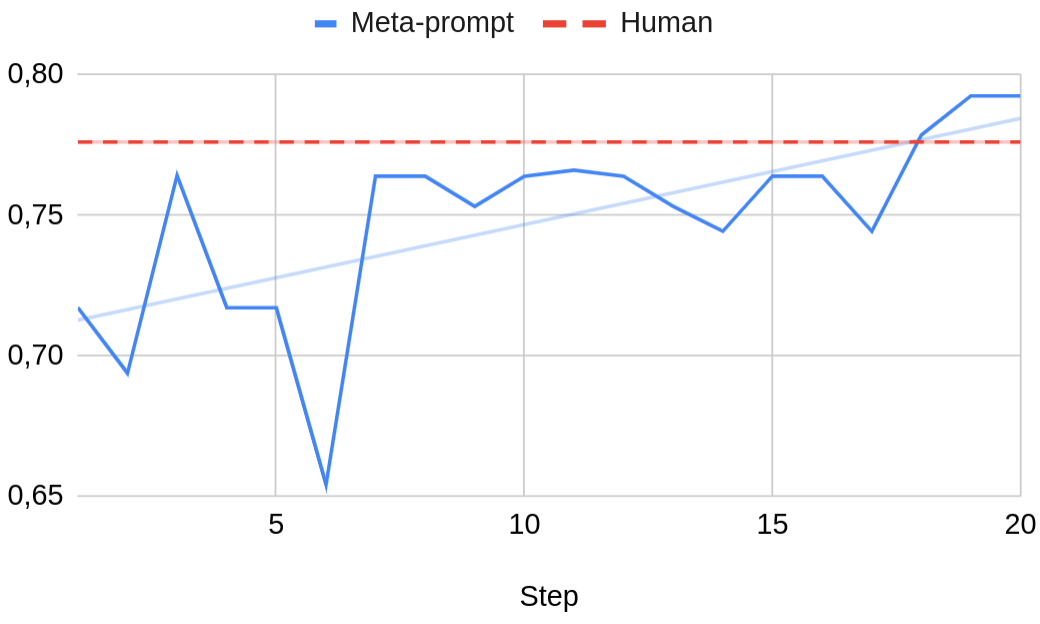}
    \caption{Performance scores with CLIP-I of the best prompt at each meta-prompting step (blue line), conducted on the first 100 examples of the test set, and the manual prompt used in the first set of experiments (red line)}
    \label{fig:graph}
\end{figure}

\begin{table}[t]
\centering
\begin{tabular}{lrrrr}
\toprule
            & BLEU              & Cosine~Sim.          & CLIP-T         & CLIP-I              \\ \midrule
Final       & 10.3595           &  \textbf{0.7695}& \textbf{0.2830}& \textbf{0.7785}     \\
Human       & \textbf{11.2302}  & 0.7611          & 0.2729         & 0.7675              \\
Initial     & 7.1965            & 0.7111          & 0.2578         & 0.7201              \\
\bottomrule
\end{tabular}
\caption{Second set of experiments, with automatic prompt and the Llama3 70B model: with final prompt (top row), initial prompt (bottom), and the manual prompt from the first set of experiments (Human)}
\label{tab:meta_prompt_results}
\end{table}

\subsection{Discussion: automatic prompt}

The scores of this second set of experiments are in Table~\ref{tab:meta_prompt_results}, while the performance of the best prompt at each optimization step is presented in Figure~\ref{fig:graph},

In Table~\ref{tab:meta_prompt_results}, one observes substantial improvement across all metrics from the initial to final step---corresponding final prompt is in Table~\ref{tab:meta_prompt_example}. This clearly indicates that the meta-prompting is an effective technique for automatically generating a prompt for our image editing method.

This is confirmed in Figure~\ref{fig:graph}, with performance, under CLIP-I, steadily trending upwards, with the manually optimized prompt outperformed at the 19th step. This optimization profile seems also to indicate that there may be room for further improvement, as performance will possibly keep growing if more meta-prompting steps are performed.

Furthermore, and above all, the scores in Table~\ref{tab:meta_prompt_results} demonstrates also that meta-prompting can lead to automatic prompts (0.2830 in CLIP-T) that outperform the prompt manually optimized in the first set of experiments (0.2729 in CLIP-T).

In what concerns the manual prompt, it is noteworthy a drop in performance with the larger 70 Billion parameter LLM (0.2729 CLIP-T, Table~\ref{tab:meta_prompt_results}), relative to the scores with the much smaller 7 Billion LLMs (0.2904 CLIP-T, Table~\ref{tab:llms}). This seems to be a certain paroxysm of the greater generative capabilities of the larger model. They may lead to longer captions for the target image, which may end up inserting ``noise'' in the editing process by enlarging the differences between the before-edit caption and the after-edit caption. 
This analysis is supported by the high BLEU and low Cosine~Sim. scores of the ``human'' prompt in Table~\ref{tab:meta_prompt_results}, given longer sequences tend to benefit BLEU metric but are irrelevant for Cosine Sim. metric. 





\end{document}